\documentclass[preprint,12pt]{elsarticle}

\usepackage{amssymb}

\usepackage{graphicx} 
\usepackage{amsmath} 
\usepackage{amssymb}  
\usepackage{bm} 
\usepackage{array}
\usepackage{colortbl}	
\usepackage[table]{xcolor}
\usepackage{subfig}

\newcommand{\degree}{\ensuremath{^\circ}}				
\DeclareMathOperator*{\argmin}{\arg\!\min}				
\DeclareMathOperator*{\st}{s.t.}						
\DeclareMathOperator*{\half}{\frac{1}{2}}					
\newcommand{\x}{\ensuremath{\times}}
\newcommand{\pinv}{\ensuremath{\dagger}}						
\newcommand{\pinvw}[2]{\ensuremath{#1^{\pinv_{#2}}}}				
\newcommand{\Rv}[1]{\ensuremath{\mathbb{R}^{#1}}}				
\newcommand{\Spd}[1]{\ensuremath{\mathbb{S}_+^{#1}}}			

\journal{Robotics and Autonomous Systems}

\begin{document}

\begin{frontmatter}



\title{Prioritized Motion-Force Control of\\ Constrained Fully-Actuated Robots:\\``Task Space Inverse Dynamics''}


\author{Andrea Del Prete$^{1,2}$, Francesco Nori$^2$, Giorgio Metta$^2$ and Lorenzo Natale$^2$}

\address{
1 --- CNRS, LAAS, 7 avenue du colonel Roche, Univ de Toulouse, \\F-31400 Toulouse, France \\
2 --- Istituto Italiano di Tecnologia, Via Morego 30, Genova, Italy
}

\begin{abstract}
We present a new framework for prioritized multi-task motion/force control of fully-actuated robots.
This work is established on a careful review and comparison of the state of the art. 
Some control frameworks are not optimal, that is they do not find the optimal solution for the secondary tasks.
Other frameworks are optimal, but they tackle the control problem at kinematic level, hence they neglect the robot dynamics and they do not allow for force control.
Still other frameworks are optimal and consider force control, but they are computationally less efficient than ours.
Our final claim is that, for fully-actuated robots, computing the operational-space inverse dynamics is equivalent to computing the inverse kinematics (at acceleration level) and then the joint-space inverse dynamics.
Thanks to this fact, our control framework can efficiently compute the optimal solution by decoupling kinematics and dynamics of the robot.
We take into account: motion and force control, soft and rigid contacts, free and constrained robots.
Tests in simulation validate our control framework, comparing it with other state-of-the-art equivalent frameworks and showing remarkable improvements in optimality and efficiency.
\end{abstract}


\end{frontmatter}

\section{Introduction}
Several frameworks for multi-task control of rigid robots exist in the literature.
Most frameworks presented in the '80s and '90s \citep{Nakamura1987, Siciliano1991, Chiaverini1997, Baerlocher1998} work at kinematic level, computing the desired joint velocities or accelerations.
These approaches are not suited to robots that interact with the environment, because they do not allow for force control or impedance control.
This motivated a more recent trend of torque control strategies \citep{Sentis2005, DeLasa2009, Saab2011, Mistry2011}, which consider the dynamics of the robot, computing the desired joint torques.
This approach can also improve tracking, because it compensates for the dynamic coupling between the joints of the multi-body system.

\citet{Peters2007} showed that we can derive several of these well-known torque control laws under a Unifying Framework (UF), as solutions of constrained minimization problems.
However, it is still unclear how these frameworks differ from each other and what their pros and cons are.
This paper has a twofold aim: first, to provide a fair comparison of the state-of-the-art torque control frameworks; second, to present a new framework which outperforms the current state of the art.
Our evaluation is based on four criteria: soundness, optimality, capabilities and efficiency.
We carry out an analytical analysis of the frameworks and we test them in simulation to confirm the theoretical results.

Section \ref{sec:key_idea} defines the basic tracking control problem and presents the main contribution of the paper through a simple example.
Section \ref{related_works} summarizes the related works and defines the notion of soundness, optimality, capabilities and efficiency.
Section \ref{sec:uf} and \ref{sec:wbcf} describe the Unifying Framework (UF) \cite{Peters2007} and the Whole-Body Control Framework (WBCF) \cite{Sentis2005}, which are the frameworks that we chose for comparison, because they well represent the state of the art.
Section \ref{sec:need} motivates the need for our new control framework Task Space Inverse Dynamics (TSID), which we then present in Section \ref{sec:tsid}.
For each framework we first discuss the solution for a single motion-control task, then we extend it to the multi-task case, and finally we introduce force control.
Section \ref{Tests} tests the three frameworks (TSID, UF, WBCF) in simulation on the same multi-task scenario, comparing their performances in terms of optimality and efficiency.
The results prove that our control framework is sound, optimal and computationally more efficient than the other frameworks.
\renewcommand{\arraystretch}{1.3}
\section{Key Idea}
\label{sec:key_idea}
\subsection{Notation and Problem Definition}
We indicate with $\Spd{n}$  the set of symmetric positive-definite $n\times n$ matrices.
We want to design position tracking control laws for a rigid manipulator with $n$ Degrees of Freedom (DoFs).
The equation of motion of a manipulator in free space may be written as \citep{Siciliano2008}:
\begin{equation}\label{man_dyn}
M(q) \ddot{q} + h(q,\dot{q}) = \tau,
\end{equation}
where $q\in \mathbb{R}^n$ are the generalized coordinates (e.g. joint angles), $\tau \in \mathbb{R}^n$ are the generalized forces (e.g. joint torques), $M(q) \in \Spd{n}$ is the joint-space mass matrix, and $h(q, \dot{q}) \in \mathbb{R}^n$ contains all the nonlinear terms such as Coriolis, centrifugal and gravity forces.
A position tracking task for the robot is described as a time-varying constraint $f(q) = x_r(t)$, where $x_r(t) \in \mathbb{R}^m$ is the reference task trajectory and $f: \Rv{n} \rightarrow \Rv{m}$ is a generic function of the generalized coordinates (e.g. the forward kinematics).
Since we assume that the control inputs are the generalized forces $\tau$, instantaneously we can only affect the generalized accelerations $\ddot{q}$.
To express the task in terms of $\ddot{q}$ we differentiate the constraint twice with respect to time:
\begin{align*}
J(q) \dot{q} &= \dot{x}_r(t), \qquad J(q) \ddot{q} + \dot{J}(q) \dot{q} = \ddot{x}_r(t),
\end{align*}
where $J(q) = \frac{\partial{}}{\partial{q}} f(q) \in \mathbb{R}^{m \times n}$ is the task Jacobian.
In the following, dependency upon $t$, $q$ and $\dot{q}$ is no longer denoted to simplify notation.
Since we use the second derivative of the constraint, in real situations a drift is likely to occur.
To prevent deviations from the desired trajectory and to ensure disturbance rejection, we design a proportional-derivative feedback control law:
\begin{equation*}
\ddot{x}^* = \ddot{x}_r + K_d (\dot{x}_r - \dot{x}) + K_p (x_r - x),
\end{equation*}
where $\ddot{x}^*\in \Rv{m}$ is the desired task acceleration, whereas $K_d \in \Spd{m}$ and \mbox{$K_p\in \Spd{m}$} are the derivative and proportional gain matrices, respectively.

\subsection{A Simple Example}
Following the approach taken in \citet{Peters2007}, we can derive task-space control laws as solutions of a constrained Quadratic Program (QP).
To better convey the idea, in this example we take a simplified form of the dynamics and kinematics:
\begin{equation*} \label{ex_ctrl_prob} \begin{split}
\tau^* = \argmin_{\tau \in \mathbb{R}^n} || \ddot{x} - \ddot{x}^* ||^2 
\quad \st \qquad & M \ddot{q} = \tau, 
\qquad J \ddot{q} = \ddot{x}
\end{split} \end{equation*}
By resolving the constraints we can transform this QP into:
\begin{equation*} \label{ex_ctrl_prob2} \begin{split}
\tau^* = \argmin_{\tau \in \mathbb{R}^n} || J M^{-1} \tau - \ddot{x}^* ||^2
\end{split} \end{equation*}
To solve this QP we can use the pseudoinverse \cite{Nakamura1987} (for the sake of simplicity here we neglect null-space terms): 
\begin{equation} \label{ex_sol}
\tau^* = \pinvw{(JM^{-1})}{} \ddot{x}^*
\end{equation}
Alternatively we can use a weighted pseudoinverse \cite{Nakamura1990}:
\begin{equation*} \label{ex_sol_weight} \begin{split}
\tau^* = \pinvw{(JM^{-1})}{W} \ddot{x}^* = W M^{-1} J^T (J M^{-1} W M^{-1} J^T)^{\pinv} \ddot{x}^*,
\end{split}\end{equation*}
where $W \in \Spd{n}$ is an arbitrary weight matrix.
The key idea that we are going to exploit is that a careful choice of $W$ can lead to a more efficient solution.
In particular, if we set $W=M^2$ we get:
\begin{equation*} \label{ex_sol_weight_M} \begin{split}
\tau^* = M J^T (J J^T)^{\pinv} \ddot{x}^* = M J^{\pinv} \ddot{x}^*
\end{split}\end{equation*}
This expression has a lower computational cost than \eqref{ex_sol}, mainly because it does not contain the inverse of the mass matrix $M$.
In general the choice of $W$ affects which value we select among the infinite solutions \cite{Nakamura1990}, so one could argue that a particular $W$ leads to a solution that is somehow \emph{better} than others \cite{Bruyninckx2000}.
However, in case of multi-task control we do not solve only one QP, but a sequence of QPs, which must have a \emph{unique solution} (see Section \ref{role_weight_mat} for details).
In this case $W$ does not affect which solution we select (because there is only one), so we claim that the choice of $W$ should be aimed only at simplifying the computational cost of the solution.

\section{Related Works}
\label{related_works}
This section provides an overview of the related works.
Table~\ref{table:frameworks} lists the main features of the control frameworks that we considered in our analysis.
\rowcolors{2}{}{gray!15}
\begin{table}[ht] 
\footnotesize
\caption{Control frameworks.}
\centering 
\begin{tabular}{p{4.1cm}  	@{}	>{\centering\arraybackslash}m{1.0cm} 	>{\centering\arraybackslash}m{1.0cm} 	>{\centering\arraybackslash}m{1.2cm}  >{\centering\arraybackslash}m{1.4cm}  		>{\centering\arraybackslash}m{1.3cm} 		>{\centering\arraybackslash}m{1cm} } 
\hline
      	 FrameworkÊÊÊÊÊÊÊÊÊÊÊÊÊÊÊÊÊÊÊÊ  & Optimal 	& Efficient		& Force Control 	& Under actuated	& Inequality	& Output 	\\ 
	 [0.5ex] \hline
ÊÊÊÊÊÊÊÊ\textbf{Task Space Inverse
	\newline Dynamics (TSID)}& $\times$ÊÊ	& $\times$ÊÊ	& {\centering $\times$}ÊÊÊÊ		& 				&			& $\tau$ÊÊÊÊÊÊÊ\\ 
ÊÊÊÊÊÊÊÊ\citet{Peters2007} (UF)	& ÊÊÊÊÊÊÊÊÊÊ 		& $\times$ÊÊÊ	& $\times$ÊÊÊ		&      			&			& $\tau$ÊÊÊÊÊÊÊ\\ 
ÊÊÊÊÊÊÊÊ\citet{Sentis2005}
	\newline(WBCF)		& $\times$	& ÊÊÊÊÊÊ 		& $\times$ÊÊÊÊÊÊ 		& ($\times$)		&			& $\tau$ÊÊÊÊÊÊÊ\\ 
ÊÊÊÊÊÊÊÊ\citet{Mistry2011}		& ÊÊÊÊÊÊÊÊÊÊ 		&   ÊÊÊ		& $\times$ÊÊÊ		& $\times$     		&			& $\tau$ÊÊÊÊÊÊÊ\\
ÊÊÊÊÊÊÊÊ\citet{Saab2011}Ê(SoT)ÊÊÊÊ	& $\times$ÊÊÊÊÊÊÊÊ&    ÊÊÊ		& $\times$ÊÊÊ		& $\times$    		&	\x		& $\tau$ÊÊÊÊÊÊÊ\\
ÊÊÊÊÊÊÊÊ\citet{DeLasa2009}ÊÊÊÊÊÊÊÊÊ	& $\times$	&   ÊÊÊ		& $\times$ÊÊÊ		& $\times$    		&			& $\tau$ÊÊÊÊÊÊÊ\\
ÊÊÊÊÊÊÊÊ\citet{Jeong2009}ÊÊÊÊÊÊÊÊÊ	& $\times$ÊÊÊÊÊÊÊÊ& $\times$ÊÊÊ	&   ÊÊÊ			&      			&			& $\tau /\ddot{q}$ÊÊÊÊÊÊÊ\\
\hline
	\citet{Smits2008} (iTaSC)	& Ê\xÊÊÊÊÊÊÊÊ 		& $\times$ÊÊÊ	&   ÊÊÊ			&      			&	\x		& $\dot{q}$ÊÊÊÊÊÊÊ\\
ÊÊÊÊÊÊÊÊ\citet{Chiaverini1997}ÊÊÊÊÊÊÊ	& ÊÊÊÊÊÊÊÊÊÊ 		& $\times$ÊÊÊ	&   ÊÊÊ			&      			&			& $\dot{q}$ÊÊÊÊÊÊÊ\\
ÊÊÊÊÊÊÊÊ\citet{Siciliano1991}ÊÊÊÊÊÊÊ	& $\times$ÊÊÊÊÊÊÊÊ& $\times$ÊÊÊ	&  Ê ÊÊ			&      			&			& $\dot{q}$ÊÊÊÊÊÊÊ\\
ÊÊÊÊÊÊÊÊ\citet{Baerlocher1998}	& Ê$\times$ÊÊÊÊÊÊÊ& $\times$ÊÊÊ	&   ÊÊÊ			&      			&			& $\dot{q}$ÊÊÊÊÊÊÊ\\
ÊÊÊÊÊÊÊÊ\citet{Nakamura1987}	& ÊÊÊÊÊÊÊÊÊÊ 		& $\times$ÊÊÊ	&   ÊÊÊ			&      			&			& $\dot{q}/\ddot{q}$ÊÊÊÊÊÊÊ\\
[0.2ex] \hline 
\end{tabular} 
\label{table:frameworks} 
\end{table}
We assess a control framework in terms of soundness, optimality, capabilities and efficiency.
Table \ref{table:frameworks} also specifies the motor commands computed by each framework (column ``Output''), which can be either joint torques $\tau$, velocities $\dot{q}$ or accelerations $\ddot{q}$.
A framework is \emph{sound} if the control action of any task does not affect the performance of any higher priority tasks.
A framework is \emph{optimal} if its control action minimizes the error of each task, under the constraint of being \emph{sound}.
The \emph{capabilities} of a framework concern the types of tasks and systems that it allows to control.
Finally, a framework is \emph{efficient} if its computational complexity is minimal, considered its capabilities (typically, the more capabilities, the higher the computational complexity).
In this context, \emph{efficiency} is strictly related to the number of computed (pseudo)inverses, matrix multiplications and the computation of the mass matrix $M$.

All the control frameworks that we analyzed are \emph{sound}.
In terms of \emph{capabilities}, Table \ref{table:frameworks} reports whether a framework allows for force control, whether it can control underactuated systems and whether it handles inequality constraints.
Since we are interested in controlling robots that interact with the environment we focus on frameworks that allow for force control.
Inequalities allow the user to define a task in terms of upper/lower bounds (e.g. joint limit avoidance, balance, visibility, and collision avoidance).
However, this feature comes at a price: the algorithm can no longer compute the solution using pseudoinverses, but it requires a QP solver.
\citet{Escande2014} reached a computation time of 1 ms on an inverse-kinematics problem --- at the price of seldom suboptimal solutions.
However, they did not consider the inverse-dynamics problem (as we do in this work), which has more than twice the number of variables and, consequently, is more computationally demanding.
In another recent work \citet{Herzoga} succeeded in controlling their robot at 1 KHz using an inverse-dynamics formulation.
Nonetheless they used a fast CPU (3.4 GHz) and the robot had only 14 DoFs; in case of more DoFs or slower CPU their method may still be too slow.
For these reasons our control framework does not include inequality constraints, even if we believe that our results could be easily generalized to handle inequalities.

The possibility to control systems that are underactuated is crucial for real-world applications, however, for the sake of conciseness, this work deals only with fully-actuated systems. 
Another paper will present our results for underactuated robots.
Besides the works cited in Table \ref{table:frameworks}, another interesting approach for underactuated robots is presented in \cite{Lee2012}.
The authors select the contact forces based on the desired rate of change of the centroidal momentum, then they find desired joint accelerations that are consistent with these contact forces and finally they compute the joint torques using a hybrid-dynamics algorithm.
\citet{Stephens2010} took a similar approach, with the main difference that they found joint accelerations and torques at the same time, resulting in a less efficient computation.

This work is motivated by the fact that no control framework that allows for force control is both optimal and efficient.
Between the five frameworks that allow for force control, we select two as representative of the state of the art and we describe them in the next sections.
Our first choice is the Unifying Framework (UF) \cite{Peters2007}, because it is the only one that allows for force control while being \emph{efficient}.
Our second choice is the Whole-Body Control Framework (WBCF) \cite{Sentis2005}, because it represents the category of ``optimal but not efficient'' frameworks (i.e. the frameworks \cite{Saab2011} and \cite{DeLasa2009}).
Even though the WBCF was extended to floating-base systems, here we consider the formulation for fully-actuated robots presented in \cite{Sentis2005} and implemented in \cite{Philippsen2011}.


\section{Unifying Framework (UF)}
\label{sec:uf}
The Unifying Framework (UF) \cite{Peters2007} formulates the control problem as a constrained minimization:
\begin{equation} \label{pos_ctrl_prob} \begin{split}
\tau^* = \argmin_{\tau \in \mathbb{R}^n} || \ddot{x} - \ddot{x}^* ||^2 \quad \st \qquad & M \ddot{q} + h = \tau \\
& J \ddot{q} + \dot{J} \dot{q} = \ddot{x}
\end{split} \end{equation}
For the typical case $m<n$, this problem has infinite solutions \cite{Nakamura1990}:
\begin{equation} \label{pos_ctrl_sol} \begin{split}
\tau^* = \pinvw{(JM^{-1})}{V} ( \ddot{x}^* - \dot{J} \dot{q} + JM^{-1}h) + (I - \pinvw{(JM^{-1})}{V} JM^{-1}) \tau_0,
\end{split}\end{equation}
where $\tau_0 \in \Rv{n}$ is an arbitrary vector, $V \in \Spd{n}$ is an arbitrary matrix and \mbox{$\pinvw{A}{V} = V^{\half} (AV^{\half})^\pinv = VA^T(AVA^T)^\pinv$} is the pseudoinverse of the matrix $A$, weighted by $V$.
If $A$ is full rank, then we can also write $\pinvw{A}{V} = VA^T(AVA^T)^{-1}$.
Choosing a particular pair $(V, \tau_0)$ we get the solution that minimizes $||V^{-\frac{1}{2}}(\tau-\tau_0)||^2$ \citep{Bjorck1996}.
Setting $\tau_0=0$ and varying $V$, we get different well-known control laws, reported in Table~\ref{tab_weight};
\rowcolors{2}{}{gray!10}
\begin{table}[htdp]
\caption{Control laws for different values of weight matrix $V$}
\centering
\begin{tabular}{c c c}
\hline
$V$ & minimize & Control law, $\tau^*$ \\
\hline
$I$ & $||\tau||^2 $ & $M^{-1}J^T(JM^{-2}J^T)^\pinv ( \ddot{x}^* - \dot{J} \dot{q} + JM^{-1}h)$ \\
$M$ & $\tau^T M^{-1} \tau$ & $J^T(JM^{-1}J^T)^\pinv ( \ddot{x}^* - \dot{J} \dot{q} + JM^{-1}h)$ \\
$M^2$ & $||M^{-1}\tau||^2 $ & $MJ^T(JJ^T)^\pinv ( \ddot{x}^* - \dot{J} \dot{q} + JM^{-1}h)$ \\
\hline
\end{tabular}
\label{tab_weight}
\end{table}
the second row reports the Operational Space control law of Khatib \citep{Khatib1987}, which selects the torques that could be generated by a hypothetical force applied at the control point.
Without loss of generality, given that $M\in \Spd{n}$, we can set $V=M^2W$, where $W\in \Spd{n}$ is another arbitrary matrix, so that \eqref{pos_ctrl_sol} simplifies to:
\begin{equation} \label{pos_ctrl_sol_simp} \begin{split}
\tau^* = M \pinvw{J}{W} ( \ddot{x}^* - \dot{J} \dot{q} + JM^{-1}h) + M N^W M^{-1} \tau_0,
\end{split}\end{equation}
where $N^W=I - \pinvw{J}{W} J$ is a weighted (nonorthogonal) null-space projector.

\subsection{Hierarchical Extension}
The Unifying Framework can manage an arbitrary number of tasks $N$, each characterized by a desired acceleration $\ddot{x}_i^*$ and a Jacobian $J_i$.
To ensure the correct management of task conflicts, the tasks need prioritization: the higher the number $i$ of the task, the higher its priority.
\begin{equation*} \begin{split}
\tau^* &= M \ddot{q}_1 \\
\ddot{q}_i &= \ddot{q}_{i+1} + N_{p(i)}^W \pinvw{J_i}{W} ( \ddot{x}_i^* - \dot{J}_i \dot{q} + J_iM^{-1}h)  \qquad i\in[1, N] \\
N_{p(i)}^W &= N_{p(i+1)}^W - \pinvw{(J_{i+1} N_{p(i+1)}^W)}{W} J_{i+1} N_{p(i+1)}^W,
\end{split} \end{equation*}
where $N_{p(i)}^W$ is a projector into the null space of all the tasks $\{j \,|\, j>i\}$, computed with the recursive formula proposed in \cite{Baerlocher1998}.
The algorithm starts by computing $\ddot{q}_N$ (using $\ddot{q}_{N+1}=0$ and $N_{p(N)}^W=I$) and it proceeds backwards up to $\ddot{q}_1$.
If the state of the robot is completely controlled, which is usually the case (see Section~\ref{role_weight_mat}), then this formulation simplifies to:
\begin{equation} \label{eq:UF}\begin{split}
\tau^* &= M \ddot{q}_1 + h \\
\ddot{q}_i &= \ddot{q}_{i+1} + N_{p(i)}^W \pinvw{J_i}{W} ( \ddot{x}_i^* - \dot{J}_i \dot{q})  \qquad i\in[1, N]
\end{split} \end{equation}
The accelerations of each task $\ddot{q}_i$ are projected into the null space of the higher priority tasks; this guarantees that the framework is \emph{sound}.
However, this approach is not \emph{optimal}, because each task is solved independently, and then projected into the null space of the higher priority tasks. 
This does not ensure the minimization of the error of each task (see \cite{Baerlocher1998, Chiaverini1997} for a thorough explanation).

\subsection{Hybrid Control}
The Unifying Framework allows for hybrid position/force control by setting the joint space control torques to:
\begin{equation*}
\tau_0 =  h - J_c^T f^*,
\end{equation*}
where $J_c(q)\in \mathbb{R}^{k \times n}$ is the contact Jacobian, $f^* \in \mathbb{R}^k$ are the desired contact forces and $k\in \mathbb{R}$ is the number of independent directions in which the robot applies force.
Substituting $\tau_0$ into the desired control torques \eqref{pos_ctrl_sol_simp} we get:
\begin{equation*}
\tau^* = M\pinvw{J}{W} ( \ddot{x}^* - \dot{J} \dot{q}) + h - M N^W M^{-1}J_c^T f^*,
\end{equation*}
where the applied forces act in the null space of the tracking task.

\section{Whole-Body Control Framework (WBCF)}
\label{sec:wbcf}
In this section we describe the WBCF presented by \citet{Sentis2005}.
This framework is based on the Operational Space Formulation \cite{Khatib1987}, which we can derive by setting $W=M^{-1}$ in \eqref{pos_ctrl_sol_simp}:
\begin{equation*}
\tau^* = J^T \underbrace{(JM^{-1}J^T)^\pinv}_{\Lambda} ( \ddot{x}^* - \dot{J} \dot{q} + JM^{-1}h) + (I-J^T \pinvw{{J^T}}{M^{-1}}) \tau_0,
\end{equation*}
where $\pinvw{J}{M^{-1}} = M^{-1}J^T \Lambda$ is the dynamically-consistent Jacobian pseudoinverse and $\Lambda$ is the task-space mass matrix.

\subsection{Hierarchical Extension}
While in case of a single task the WBCF and the UF are equivalent, their hierarchical extensions differ substantially:
\begin{equation} \label{eq:WBCF} \begin{split}
\tau^* &= \sum_{i=1}^N J_{p(i)}^T F_{p(i)} \\
F_{p(i)} &= \Lambda_{p(i)} ( \ddot{x}_i^* - \dot{J}_i \dot{q} + J_iM^{-1}(h - \sum_{j=1}^{i-1} J_{p(j)}^T F_{p(j)})) \\
J_{p(i)} &= J_i (I - \sum_{j=1}^{i-1} J_{p(j)}^{\pinv_{M^{-1}}} J_{p(j)})
\end{split} \end{equation}
This prioritization strategy is different with respect to \eqref{eq:UF}: the WBCF minimizes the error of each task under the constraint of not conflicting with any higher priority tasks, namely it is \emph{optimal}.
However, this formulation is computationally less efficient than \eqref{eq:UF} because i) it contains the term $M^{-1}$ and ii) it requires more matrix multiplications.

\subsection{Hybrid Control}
The WBCF allows for hybrid position/force control by setting:
\begin{equation*}
F_{p(i)} = \Omega_f f_i^* + \Lambda_{p(i)} ( \Omega_m \ddot{x}_i^* - \dot{J}_i \dot{q} + J_iM^{-1}(h - \sum_{j=1}^{i-1} J_{p(j)}^T F_{p(j)})) ,
\end{equation*}
where the selection matrices $\Omega_f$ and $\Omega_m$ split the control space into force and motion components, respectively.


\section{The Need for a New Control Framework}
\label{sec:need}
The WBCF is \emph{sound} and \emph{optimal}, but it is not \emph{efficient} because it requires the computation of the \emph{operational space inertia matrices} $\Lambda$'s. 
The simplest way to compute them is using the formula $\Lambda = (JM^{-1}J^T)^\pinv$, which has a complexity of O($n^3$):
the computation of $M$ --- with Recursive Newton-Euler Algorithm (RNEA) or Composite-Rigid-Body algorithm \citep{Siciliano2008} --- has a complexity of O($n^2$) for serial robots and O($nd$) for multi-branch robots (where $d$ is the tree depth).
More efficient algorithms \citep{Khatib2000} can compute $\Lambda$ with a complexity of O($nm^2+m^3$), where $m$ is the dimension of the task.

On the other hand the UF is \emph{sound} and \emph{efficient} (if we choose $V=M^2$, i.e. $W=I$): the solution takes the form $\tau^* = M\ddot{q}_1 + h$, which we can calculate without explicitly computing $M$, through the O($n$) RNEA.
Nonetheless, the UF is not \emph{optimal}: even if a task does not conflict with any higher priority tasks, it may not be performed correctly.

The derivation of our framework, TSID, follows the same principles underlying the UF, but with a different hierarchical extension.
We minimize the error of each task under the constraint of not affecting any higher priority task.
At each minimization step, we carefully select the weight matrices used in the pseudoinverses, so as to simplify the resulting control laws.
This leads to an efficient formulation, while preserving the optimality property.
We start considering position tracking control only, then we introduce force control tasks.

\subsection{Weight Matrix and Joint Space Stabilization}
\label{role_weight_mat}
The weight matrix $V$ (or equivalently $W$) introduced in the resolution of \eqref{pos_ctrl_prob} can play two different roles.
In case there is no secondary task (i.e. $\tau_0=0$), $V$ determines the quantity that we minimize (e.g. $||\tau||^2$, $||\ddot{q}||^2$, $||M^{-\frac{1}{2}}\tau||^2$).
In case there is a secondary task, $V$ specifies the metric that is used to measure the distance between $\tau$ and $\tau_0$.

Using the null space of a task to minimize some measure of effort is appealing, mainly because it is rooted in the study of human motion \citep{Hogans1985}.
This approach may be feasible in simulation, but unfortunately in reality it leads to singular configurations and hitting of joint limits \citep{Peters2007}.
The subspace of joint accelerations that does not affect the task is not controlled, so its behavior is determined by disturbances and errors in the model of the manipulator.
Even in simulation, if the robot has nonzero joint velocities when the controller starts, failing to use a secondary task may result in joint space instability.
The reason for this behavior is obvious: the effort of stabilizing in joint space is not task relevant and it would increase the cost \citep{Peters2007}.

\citet{Peters2007} suggest to add a joint space motor command for stabilization.
A common approach is to design the postural task to attract the robot towards a desired posture $q_p$.
We compute the desired joint accelerations as $\ddot{q}_p^*=K_p(q_p-q)-K_d\dot{q}$, where $K_p\in \Spd{n}$ and $K_d\in \Spd{n}$ are the proportional and damping gain matrices.
In the following we always include the postural task to minimize $|| \ddot{q}-\ddot{q}_p^* ||^2$, under the constraint of not affecting any other task.
This ensures stabilization of the manipulator in joint space.

\section{Original Contribution - Task Space Inverse Dynamics (TSID)}
\label{sec:tsid}
In this section we derive the TSID control framework, which is the main contribution of the paper.
The TSID is \emph{sound}, \emph{optimal}, \emph{efficient} --- as confirmed by the simulation tests --- and allows for both motion and force control.

\subsection{Framework Derivation}
Consider a general scenario in which the robot has to perform $N$ position tracking tasks $T_1$\dots$T_N$ and a postural task $T_0$ (with desired joint accelerations $\ddot{q}_p^*$) to stabilize any left redundancy.
Taking inspiration from the UF and from \cite{DeLasa2009} we formulate the control problem as a sequence of constrained minimization, starting from the highest-priority task $N$ and moving down to the lowest-priority task $0$ (i.e. the postural task):
\begin{equation} \label{eq:tsid_prob} \begin{aligned}
(T_N) \quad & g_N^* &=& \min_{\tau \in \mathbb{R}^n} g_N (\tau) & \st \quad & M\ddot{q}+h=\tau & \\
(T_i) \quad & g_i^* &=& \min_{\tau \in \mathbb{R}^n} g_i (\tau) & \st \quad & M\ddot{q}+h=\tau, \quad g_j(\tau) = g_j^* \quad \forall j>i \\
(T_0) \quad & \tau^* &=& \argmin_{\tau \in \mathbb{R}^n} || \ddot{q}-\ddot{q}_p^* || & \st \quad & M\ddot{q}+h=\tau, \quad g_j(\tau) = g_j^* \quad \forall j>0,\\
\end{aligned}\end{equation}
where $g_i(\tau) = || J_i \ddot{q} + \dot{J}_i \dot{q} - \ddot{x}_i^* ||^2$ is the cost associated to the task $T_i$.
The solution of \eqref{eq:tsid_prob} is given by:
\begin{equation} \label{hierarchical_formulation}\begin{split}
\tau^* =& M \ddot{q}_0 + h \\
\ddot{q}_i =& \ddot{q}_{i+1} + N_{p(i)}^W \pinvw{(J_i N_{p(i)}^W)}{W} (\ddot{x}_i^*-\dot{J}_i\dot{q} - J_i\ddot{q}_{i+1}) \qquad i \in [0, N],
\end{split} \end{equation}
where $J_0=I$ and $\ddot{x}_0^* = \ddot{q}_p^*$.
The computation is initialized setting $\ddot{q}_{N+1}=0$ and $N_{p(N)}=I$.
Once again, selecting the weight matrix $W$ we can vary the form of the control law.
Interestingly enough though, the solution $\tau^*$ is independent of $W$. 
This is because the only role of $W$ is to weight the quantity that is minimized in the null space of all the tasks, but here the postural task ensures that there is no null space left (because any control action affects the postural task).
It is then reasonable to choose $W$ so as to simplify the computation. 
If we set $W=I$ then all the null-space projectors $N_{p(i)}$ become orthogonal, so they are equal to their pseudoinverses (i.e. $N_{p(i)}^\pinv=N_{p(i)}$).
This simplifies the formulation \eqref{hierarchical_formulation} to:
\begin{equation} \label{hierarchical_formulation_simp} \begin{split}
\tau^* =& M(\ddot{q}_1 + N_{p(0)}\ddot{q}_p^*) + h\\
\ddot{q}_i =& \ddot{q}_{i+1} + (J_i N_{p(i)})^\pinv (\ddot{x}_i^*-\dot{J}_i\dot{q} - J_i\ddot{q}_{i+1}) \qquad i\in[1, N]
\end{split}\end{equation}
In this form, kinematics and dynamics are completely decoupled: first we solve the multi-task prioritization at kinematic level computing $\ddot{q}_1$, then we compute the torques to get the desired joint accelerations.
This formulation does not require the computation of a pseudoinverse for the postural task, because it exploits the property of orthogonal projectors of being equal to their pseudoinverses.
Moreover, it can be efficiently computed with the RNEA, without explicitly calculating $M$.

\subsection{Force Control}
This subsection extends TSID to force control.
If the manipulator is in contact with the environment, its equations of motion become:
\begin{equation} \label{man_dyn_cont}
M(q) \ddot{q} + h(q,\dot{q}) - J_c(q)^T f = \tau,
\end{equation}
where $J_c(q) = \frac{\partial{x_c}}{\partial{q}} \in \mathbb{R}^{k \times n}$ is the contact Jacobian (or constraint Jacobian), $x_c \in \mathbb{R}^k$ is the robot contact point and $f \in \mathbb{R}^k$ are the contact forces (or constraint forces).
To control the contact forces we need a model of the contact dynamics.
The most common choices are the \emph{linear-spring contact} model \cite{Park2008} and the \emph{rigid contact} model.
The first model assumes that the environment at the contact point behaves like a linear spring, i.e. \mbox{$k_s (x_c - x_e) = f$}, where $k_s$ is the contact stiffness and $x_e \in \mathbb{R}^k$ is the environment contact point.
Assuming $k_s$ is known, force is a known function of position, so we can easily translate this kind of force control problems into position control problems.

More interesting is instead the \emph{rigid contact} model, mainly because it introduces constraints into the problem formulation.
When the manipulator is in rigid contact with the environment, its motion is subject to $k$ nonlinear constraints.
In general we can consider these constraints as nonlinear function of the generalized coordinates, their derivatives and time: $c(q, \dot{q}, t) = 0$ \footnote{The constraints may be time-varying, hence we can model contacts with curved surfaces, as long as $c$ is sufficiently smooth.}.
To include these constraints into the control problem we express them at acceleration level\footnote{In case of nonholonomic constraints we differentiate them once, whereas in case of holonomic constraints we differentiate them twice. For instance, in case of time-invariant rigid contacts we have: $b(q, \dot{q}) = - \dot{J}_c(q,\dot{q}) \dot{q}$.} as: $J_c(q) \ddot{q} = b(q,\dot{q},t)$. 
We write then the problem as:
\begin{equation} \label{eq:rig_force_prob}\begin{split}
\tau^* = \argmin_{\tau \in \mathbb{R}^n} || f - f^* ||^2 \quad \st \qquad& M \ddot{q} + h -J_c^T f = \tau \\
& J_c \ddot{q} = b,
\end{split} \end{equation}
where $f^* \in \mathbb{R}^k$ are the desired contact forces.
We can express the infinite solutions of the problem \eqref{eq:rig_force_prob} as:
\begin{equation} \label{eq:rig_force_prob_sol}
\tau^* = M (J_c^\pinv b + N_c \ddot{q}_0) + h - J_c^T f^*,
\end{equation}
where $\ddot{q}_0\in \Rv{n}$ is an arbitrary vector.
It is trivial to show that \eqref{eq:rig_force_prob_sol} is a solution of \eqref{eq:rig_force_prob} because it respects the constraints and it results in the minimum cost (i.e. $0$).
This control law is one of our main contributions, because it allows to implement force control without computing $M$, while characterizing the redundancy of the task through $\ddot{q}_0$.

\subsection{Integration of Force Control in Hierarchical Framework}
We extend the multi-task formulation \eqref{hierarchical_formulation_simp} to include force control tasks.
The rigid force control task, if any, has to take the highest priority because it is a physical constraint that cannot be violated by definition.
We assume that the robot has to perform $N-1$ position control tasks. 
On top of that there is a rigid force control task $N$ (for the sake of simplicity, here we assume holonomic constraints, i.e. $b(q,\dot{q},t) = -\dot{J}_c \dot{q}$), with reference force $f^*$ and Jacobian $J_N=J_c$:
\begin{equation} \label{hierarchical_formulation_force} \begin{split}
\tau^* =& M(\ddot{q}_1 + N_{p(0)}\ddot{q}_p^*) + h - J_c^Tf^*\\
\ddot{q}_i =& \ddot{q}_{i+1} + (J_i N_{p(i)})^\pinv (\ddot{x}_i^*-\dot{J}_i\dot{q} - J_i\ddot{q}_{i+1}) \qquad i\in [1, N],
\end{split}\end{equation}
where $\ddot{x}_N^*=\ddot{x}_c=0$, $\ddot{q}_{N+1}=0$, and $N_{p(N)}=I$.
Even after the extension to force control, kinematics and dynamics are still decoupled, so the computational complexity has not increased and $\tau^*$ can be efficiently computed with the RNEA.
\section{Tests}
\label{Tests}

\subsection{Simulation Environment}
We tested our control framework --- Task Space Inverse Dynamics --- against the Unifying Framework (UF) \cite{Peters2007} and the Whole-Body Control Framework (WBCF) \cite{Sentis2005},
on a customized version of the Compliant huManoid (CoMan) simulator \cite{Dallali}.
The robot has 23 DoFs: 4 in each arm, 3 in the torso and 6 in each leg.
We adapted the simulator to make the robot rigid and fully-actuated (we fixed the robot base and we removed the joint passive compliance).
Direct and inverse dynamics, both in simulation and control, were efficiently computed using C language functions, generated with the Robotran \cite{Robotranwebpage} symbolic engine.
Contact forces were simulated using linear spring-damper models (stiffness $2 \cdot10^5 N/m$ and damping $10^3 Ns/m$, as proposed in \cite{Dallali}) with realistic friction.
To integrate the equations of motion we used the Simulink variable step integrator \emph{ode23t}, with relative and absolute tolerance of $10^{-3}$ and $10^{-6}$, respectively.
The tests were executed on a computer with a 2.83 GHz CPU and 4 GB of RAM.
The computation times are computed as averages over the whole test (i.e. some thousands of executions).

\subsection{Trajectory Generation}
To generate reference position-velocity-acceleration trajectories we used the approach presented in \cite{Pattacini2010}, which provides approximately minimum-jerk trajectories.
The trajectory generator is a 3rd order dynamical system that takes as input the desired trajectory $x_d(t)$ and outputs the three position-velocity-acceleration reference trajectories $x_r(t), \dot{x}_r(t), \ddot{x}_r(t)$.
The reference position trajectory follows the desired position trajectory with a velocity that depends on the parameter ``trajectory time'' (always set to $1.0 s$ in our tests).
We set all proportional gains $K_p=10 s^{-2}$, and all derivative gains $K_d=5 s^{-1}$.

\subsection{Damped Pseudoinverses}
\label{subsec:damp_pinv}
The controllers used damped pseudoinverses \cite{Chiaverini1997} to ensure stability near singularities.
Based on our experience on a real robot, we set the damping factor $\lambda = 0.02$, which ensures a maximum gain of the pseudoinverses of $(2\lambda)^{-1} = 25$.
To avoid interferences between tasks, null-space projection matrices were computed without any damping, but setting to zero all singular values below the threshold $\sigma_{min} = 2.5\cdot10^{-8}$.
The values $\lambda$ and $\sigma_{min}$ must be chosen so that: \mbox{$\sigma_{min}(\sigma_{min}^2+\lambda^2)^{-1} < z$}, where $z$ is a small positive value (e.g. we chose $z\simeq10^{-4}$).
This ensures coherence between damped pseudoinverses and null-space projectors, so that any direction of the control space is not used by more tasks at the same time.
The use of damped pseudoinverses modifies the minimization problem \eqref{eq:tsid_prob}, adding a regularization term $\lambda ||\tau||^2_W$ to the cost functions.
Away from singularities this term has negligible effects, but close to singularities it keeps $||\tau^*||$ bounded, at the expenses of the task errors.
If $\lambda=0$, we know that WBCF and TSID give the same results. 
However, with $\lambda \neq 0$, since the regularization term is affected by the weight matrix $W$, we expect to see some differences between WBCF and TSID when close to singularities.

\subsection{Test 1 - Feasible Task Hierarchy}
\begin{figure}[htbp]
\centering
\subfloat[]{
\parbox{0.3\textwidth}{\includegraphics[width=0.25\textwidth]{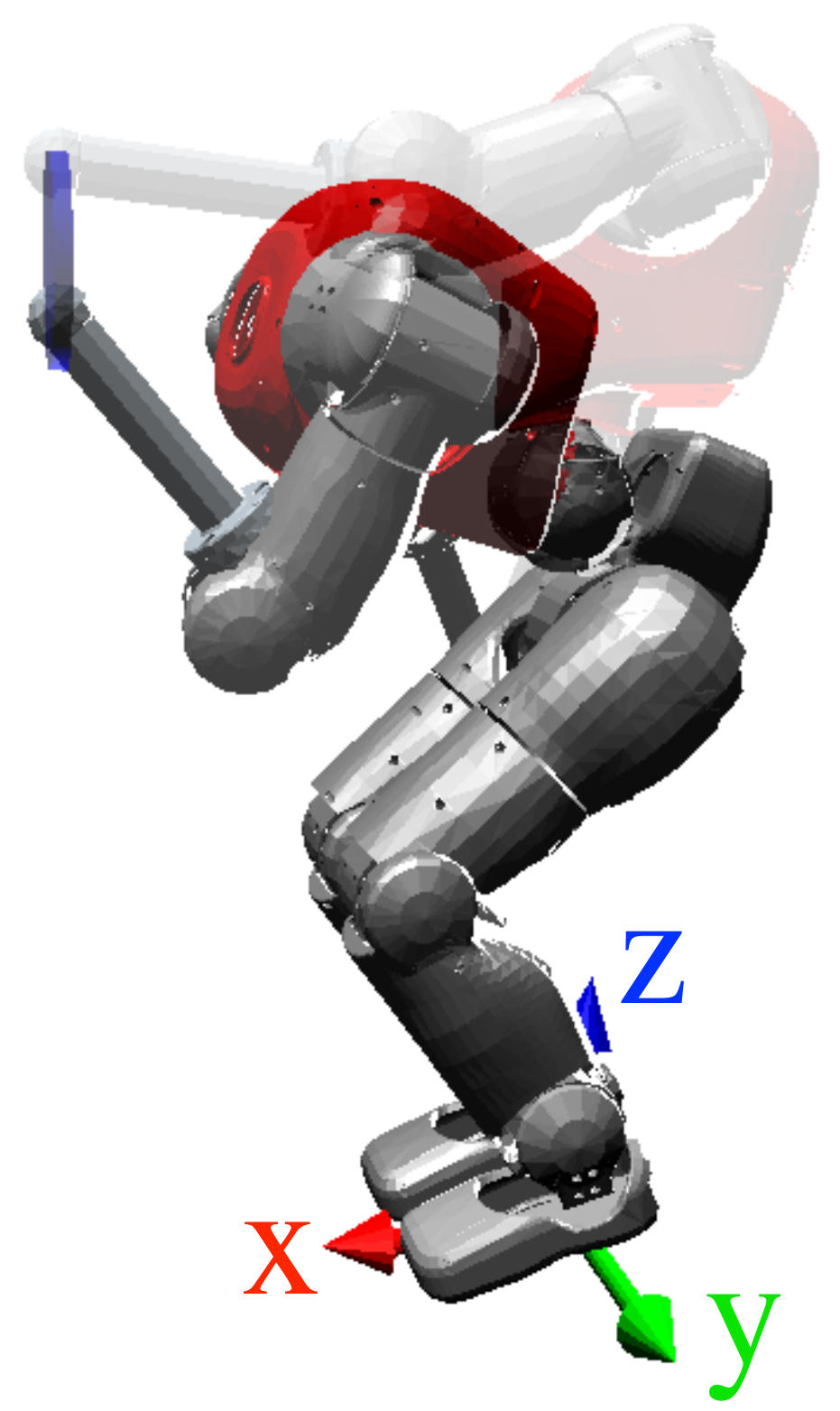}}%
\label{coman_circle}}%
\quad%
\subfloat[]{%
\parbox{0.4\textwidth}{\includegraphics[width=0.35\textwidth]{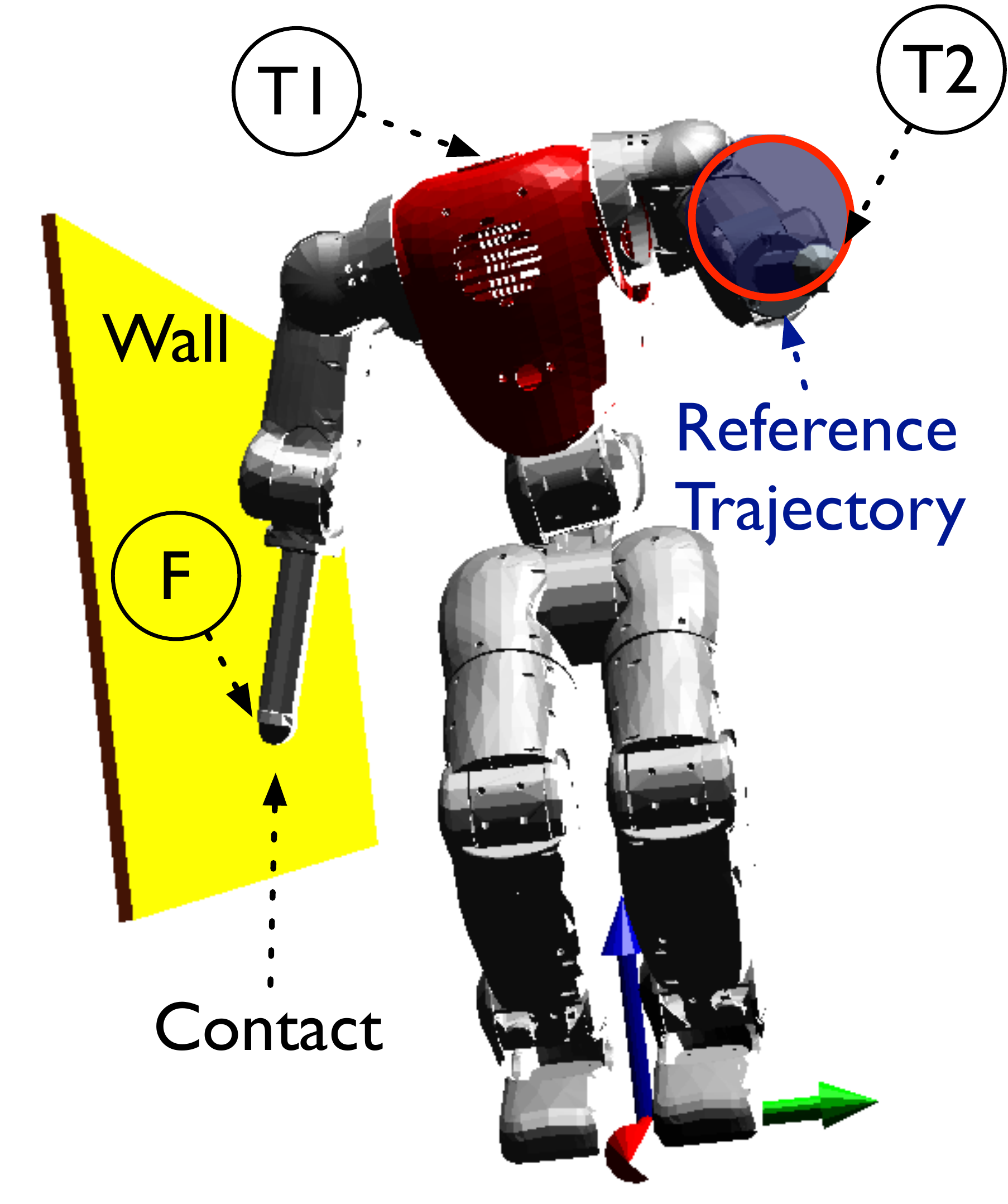}}%
\label{coman_circle2}}%
\caption{ \small CoMan \cite{Dallali} executing Test 1. Task $F$ controls the force exerted by the right hand against the wall. Task $T_2$ moves the left hand along the circular reference trajectory depicted as a red circumference. Task $T_1$ moves the neck base back and forth along the x axis.}
\end{figure}
In this test the robot performs four tasks:
\begin{description}
\item[$F$:] 3 DoFs, apply a normal force of 20 N on the wall with the right hand
\item[$T_2$:] 3 DoFs, track a circular trajectory with the left hand
\item[$T_1$:] 1 DoF (x coordinate), track a sinusoidal reference with the neck base
\item[$T_0$:] 23 DoFs, maintain the initial joint posture
\end{description}
The first three tasks are always compatible, so the robot should be able to perform them with negligible errors.
Table \ref{table:test1_res} reports the root-mean-square error (RMSE) for each task and the mean computation time of the control loop.
We compute the RMSE as $\sqrt{\frac{1}{N_t} \sum_{t=0}^{T} || x(t) - x_r(t) ||^2}$, where $N_t$ is the number of samples used in the summation.
The criteria proposed in Section \ref{related_works} (in particular see Table~\ref{table:frameworks}) are strictly connected to the data of Table \ref{table:test1_res}: the error of the primary task F concerns the \emph{soundness}, the errors of the nonprimary tasks ($T_2$, $T_1$, $T_0$) concern the \emph{optimality}, and the computation time concerns the \emph{efficiency}.
\rowcolors{4}{}{gray!15}
\begin{table}[ht] 
\small
\caption{\small Test 1. Root-mean-square error of the four tasks and average computation time of the controller.}
\centering 
\begin{tabular}{l c c c c c} 
	Related to 			& \cellcolor[gray]{0.8} \emph{Soundness} & \multicolumn{3}{c}{\emph{Optimality}} \cellcolor[gray]{0.8} & \emph{Efficiency} \\
\hline
      	 ControllerÊÊÊÊÊÊÊÊÊÊÊÊÊÊÊÊÊÊÊÊÊÊÊ & F-RMSE			& $T_2$-RMSE 		& $T_1$-RMSE  	& $T_0$-RMSE	& Computation		\\ 
      	 		ÊÊÊÊÊÊÊÊÊÊÊÊÊÊÊÊÊÊÊÊÊÊÊ &  (N)			&  (mm) 			& (mm) 			& (\degree)		& Time (ms)		\\ 
	 [0.3ex] \hline
ÊÊÊÊÊÊÊÊ\textbf{TSID}			& \textbf{0.1} 		& \textbf{0.4}ÊÊÊ 		& \textbf{0.1}ÊÊÊ 		& \textbf{7.1}		& \textbf{0.24}ÊÊÊÊÊÊÊ\\ 
ÊÊÊÊÊÊÊÊWBCFÊÊÊÊÊÊÊÊÊÊÊ			& 0.1 ÊÊÊÊÊÊÊÊÊÊ 		& 0.4ÊÊÊÊÊÊ 			& 0.1ÊÊÊÊÊÊ 			& 7.1			& 0.64ÊÊÊÊÊÊÊ		\\ 
ÊÊÊÊÊÊÊÊUFÊÊÊÊÊÊÊÊÊ				& 0.1 ÊÊÊÊÊÊÊÊÊÊ 		& 36.8ÊÊÊ			& 30.1ÊÊÊ			& 6.6			& 0.25ÊÊÊÊÊÊÊ		\\ 
[0.1ex] \hline 
\end{tabular} 
\label{table:test1_res} 
\end{table}
As expected, the UF performs poorly on the nonprimary tasks, because it is not optimal.
Both WBCF and TSID achieve good tracking on all tasks, but the computation time of WBCF is $\sim2.6\times$ the computation time of our framework.

\subsection{Test 2 - Unfeasible Task Hierarchy}
In this test the robot performs the same four tasks of the previous test, with the only difference that task $T_1$ controls the 3D Cartesian position of the neck base (rather than the x coordinate only).
\rowcolors{4}{}{gray!15}
\begin{table}[ht] 
\small
\caption{\small Test 2. Root-mean-square error of the four tasks and average computation time of the controller.}
\centering 
\begin{tabular}{l c c c c c} 
	Related 			& \cellcolor[gray]{0.8} \emph{Soundness} & \multicolumn{3}{c}{\emph{Optimality}} \cellcolor[gray]{0.8} & \emph{Efficiency} \\
\hline
      	 ControllerÊÊÊÊÊÊÊÊÊÊÊÊÊÊÊÊÊÊÊÊÊÊÊ & F-RMSE			& $T_2$-RMSE 		& $T_1$-RMSE  	& $T_0$-RMSE		& Computation		\\ 
      	 		ÊÊÊÊÊÊÊÊÊÊÊÊÊÊÊÊÊÊÊÊÊÊÊ &  (N)			&  (mm) 			& (mm) 			& (\degree)		& Time (ms)		\\ 
	 [0.3ex] \hline
ÊÊÊÊÊÊÊÊ\textbf{TSID}			& \textbf{0.0} 		& \textbf{0.1}ÊÊÊ 		& \textbf{21.5}Ê 		& \textbf{5.5}		& \textbf{0.25}ÊÊÊÊ   ÊÊÊ	\\ 
ÊÊÊÊÊÊÊÊWBCFÊÊÊÊÊÊÊÊÊÊÊ			& 0.0 ÊÊÊÊÊÊÊÊÊÊ 		& 0.3ÊÊÊÊÊÊ 			& 21.5ÊÊÊÊÊ 			& 5.5			& 0.67ÊÊÊÊÊÊÊ		\\ 
ÊÊÊÊÊÊÊÊUFÊÊÊÊÊÊÊÊÊ				& 0.0 ÊÊÊÊÊÊÊÊÊÊ 		& 23.8ÊÊÊ			& 62.4ÊÊÊ			& 5.1			& 0.26ÊÊÊÊÊÊÊ		\\ 
[0.1ex] \hline 
\end{tabular} 
\label{table:test2_res} 
\end{table}
\rowcolors{1}{}{}
This makes impossible to achieve all tasks at the same time (the desired neck trajectory is not reachable), so we expect a significant error for task $T_1$, while the tasks $F$ and $T_2$ should have negligible errors.
Table \ref{table:test2_res} shows that, as before, UF performs poorly on the nonprimary tasks, whereas WBCF has higher computation time than the other two frameworks.
The small difference between TSID and WBCF in the RMSE of task $T_2$ is due to the behavior of the damped pseudoinverses when close to the singularity due to the conflict between task $T_1$ and the tasks $F$ and $T_2$.
While this effect is clear from a numerical standpoint (see Section \ref{subsec:damp_pinv}), there seems not to be a \emph{best} choice for the weight to use for damping; indeed we observed that, when close to singularities, sometimes WBCF performs slightly worse than TSID, but other times it performs slightly better.


\section{Conclusions}
\label{Conclusions}
We presented and validated a new theoretical control framework, called Task Space Inverse Dynamics, for prioritized motion and force control of fully-actuated robots.
To the best of our knowledge, this framework outperforms every other control framework with equal capabilities.
Its main features are:
\begin{enumerate}
\item \emph{optimality}: it minimizes the error of each task under the constraint of not affecting any higher priority task
\item \emph{capabilities}: it allows for position/velocity/acceleration control and soft/rigid contact force control
\item \emph{efficiency}: it computes the desired joint torques in O($n$) using the Recursive Newton-Euler Algorithm because it needs neither the joint-space mass matrix $M$, nor the task-space mass matrices $\Lambda$'s
\end{enumerate}
We compared the presented control framework with other two state-of-the-art control frameworks (UF and WBCF), both analytically and through simulation tests.
We decided to carry out this comparison in simulation to avoid that model inaccuracies could interfere with the controllers in unpredictable ways.
The results confirm that our framework outperforms the other two frameworks, either in terms of optimality or efficiency.

Moreover this work proves that, for fully-actuated robots, it is not necessary to take into account the dynamics when resolving the multi-task motion/force control problem.
In other words, the WBCF is equivalent to a second-order inverse kinematics --- which computes the desired joint accelerations --- followed by an inverse dynamics --- which computes the desired joint torques.

\subsection{Discussion and Future Work}
The presented framework suffers from three main limitations, which we should address in practical applications.
\begin{enumerate}
\item It does not allow for inequality constraints \cite{Saab2011, Smits2008}, which are particularly important for modeling joint limits and motor torque bounds.
\item It does not deal with planning and it guarantees only instantaneous (local) optimality, so a task may lead the robot into a configuration in which a higher priority task becomes singular, hence unfeasible.
To tackle this issue the cost function should include not only instantaneous errors, but the summation of errors over a certain time horizon (i.e. model predictive control, MPC \citep{Manchester2011, Tassa2012}).
\item We only considered fully-actuated robots, while many mechanical systems are underactuated (e.g. floating-base, underwater, elastic robots).
\end{enumerate}
While these limitations could make this contribution seem negligible, we argue the opposite.
The frameworks that tackle the limitations 1 and 2 typically work by iteratively solving simplified control problems such as the ones we discussed (i.e. nonlinear functions are linearized around the current solution, whereas inequality constraints are converted into equality constraints using active-set methods).
This implies that more advanced frameworks could exploit the presented results to improve their efficiency --- which often is the bottleneck preventing their applications on real robots \citep{Tassa2012}.
Regarding the limitation 3, for the sake of conciseness this work has dealt only with fully-actuated mechanical systems.
A future paper will present an extension of this framework for floating-base robots, which relies on the same principles and techniques that we used here.
Finally, we are now in the process of testing TSID on a real humanoid robot.

\section*{Acknowledgement}
This paper was supported by the FP7 EU projects CoDyCo (No. 600716 ICT 2011.2.1 Cognitive Systems and Robotics), and Koroibot (No. 611909 ICT-2013.2.1 Cognitive Systems and Robotics).




\bibliographystyle{model1-num-names}
\bibliography{TaskInvDynCtrl}







\end{document}